\documentclass[runningheads]{llncs}
\usepackage{graphicx}

\usepackage{tikz}
\usepackage{comment} 
\usepackage{amsmath,amssymb} 
\usepackage{color}
\usepackage{url}
\usepackage{hyperref}
\hypersetup{
    colorlinks=true,
    linkcolor=blue,
    filecolor=blue,      
    urlcolor=magenta,
    citecolor=blue,
}

\usepackage{multirow}

\begin{document}
\pagestyle{headings}
\mainmatter
\def\ECCVSubNumber{986}  

\title{Linguistic Structure Guided Context Modeling for Referring Image Segmentation} 

\titlerunning{LSCM for Referring Image Segmentation}
%
\author{Tianrui Hui\inst{1,2} \and
Si Liu\inst{3}\thanks{Corresponding author} \and
Shaofei Huang\inst{1,2} \and
Guanbin Li\inst{4} \and
Sansi Yu\inst{5} \and \\
Faxi Zhang\inst{5} \and
Jizhong Han\inst{1,2}}
\authorrunning{T. Hui et al.}
%
\institute{Institute of Information Engineering, Chinese Academy of Sciences \and
School of Cyber Security, University of Chinese Academy of Sciences \and
Institute of Artificial Intelligence, Beihang University \and
Sun Yat-sen University \and
Tencent Marketing Solution \\
\email{\{huitianrui,huangshaofei,hanjizhong\}@iie.ac.cn}; \\
\email{liusi@buaa.edu.cn}; \email{liguanbin@mail.sysu.edu.cn}; \\
\email{\{mionyu,micahzhang\}@tencent.com}}
\maketitle

\begin{abstract}
Referring image segmentation aims to predict the foreground mask of the object referred by a natural language sentence. 
Multimodal context of the sentence is crucial to distinguish the referent from the background. 
Existing methods either insufficiently or redundantly model the multimodal context.
To tackle this problem, we propose a ``gather-propagate-distribute'' scheme to model multimodal context by cross-modal interaction and implement this scheme as a novel Linguistic Structure guided Context Modeling (LSCM) module. 
Our LSCM module builds a Dependency Parsing Tree suppressed Word Graph (DPT-WG) which guides all the words to include valid multimodal context of the sentence while excluding disturbing ones through three steps over the multimodal feature, i.e., gathering, constrained propagation and distributing. 
Extensive experiments on four benchmarks demonstrate that our method outperforms all the previous state-of-the-arts.
Code is available at \url{https://github.com/spyflying/LSCM-Refseg}.
\end{abstract}

\section{Introduction}
Referring image segmentation aims at predicting the foreground mask of the object which is matched with the description of 
a natural language expression. It enjoys a wide range of applications, e.g., human-computer interaction and interactive 
image editing. 
Since natural language expressions may contain diverse linguistic concepts, 
such as entities (e.g. ``car'', ``man''), attributes (e.g. ``red'', ``small'') and relationships (e.g. ``front'', ``left''), 
this task is faced with a broader set of categories compared with a predefined one in traditional semantic segmentation. 
It requires the algorithm to handle the alignment of different semantic concepts between language and vision. 

A general solution to this task is first extracting visual and linguistic features respectively, and then conducting segmentation based on the multimodal features generated from the two types of features. 
The entity referred by a sentence is defined as the \textit{referent}. 
Multimodal features of the referent is hard to be distinguished from features of the background due to the existence of abundant noises. 
To solve this problem, valid multimodal context relevant to the sentence can be exploited to highlight features of the referent and suppress those of the background for accurate segmentation. 
Some works tackle this problem by straightforward concatenation~\cite{hu2016segmentation}\cite{shi2018key} or recurrent refinement~\cite{liu2017recurrent}\cite{li2018referring}\cite{chen2019see} of visual and linguistic features but lack the explicit modeling of multimodal context. 
Other works introduce dynamic filters~\cite{margffoy2018dynamic} or cross-modal self-attention~\cite{ye2019cross} to model multimodal context. However, these multimodal contexts are either insufficient or redundant since the number of dynamic filters~\cite{margffoy2018dynamic} is limited and weights for aggregating multimodal context in self-attention~\cite{ye2019cross} may be redundant due to dense computation operations.

\begin{figure}[t]
	\centering
	\includegraphics[width=0.52\linewidth]{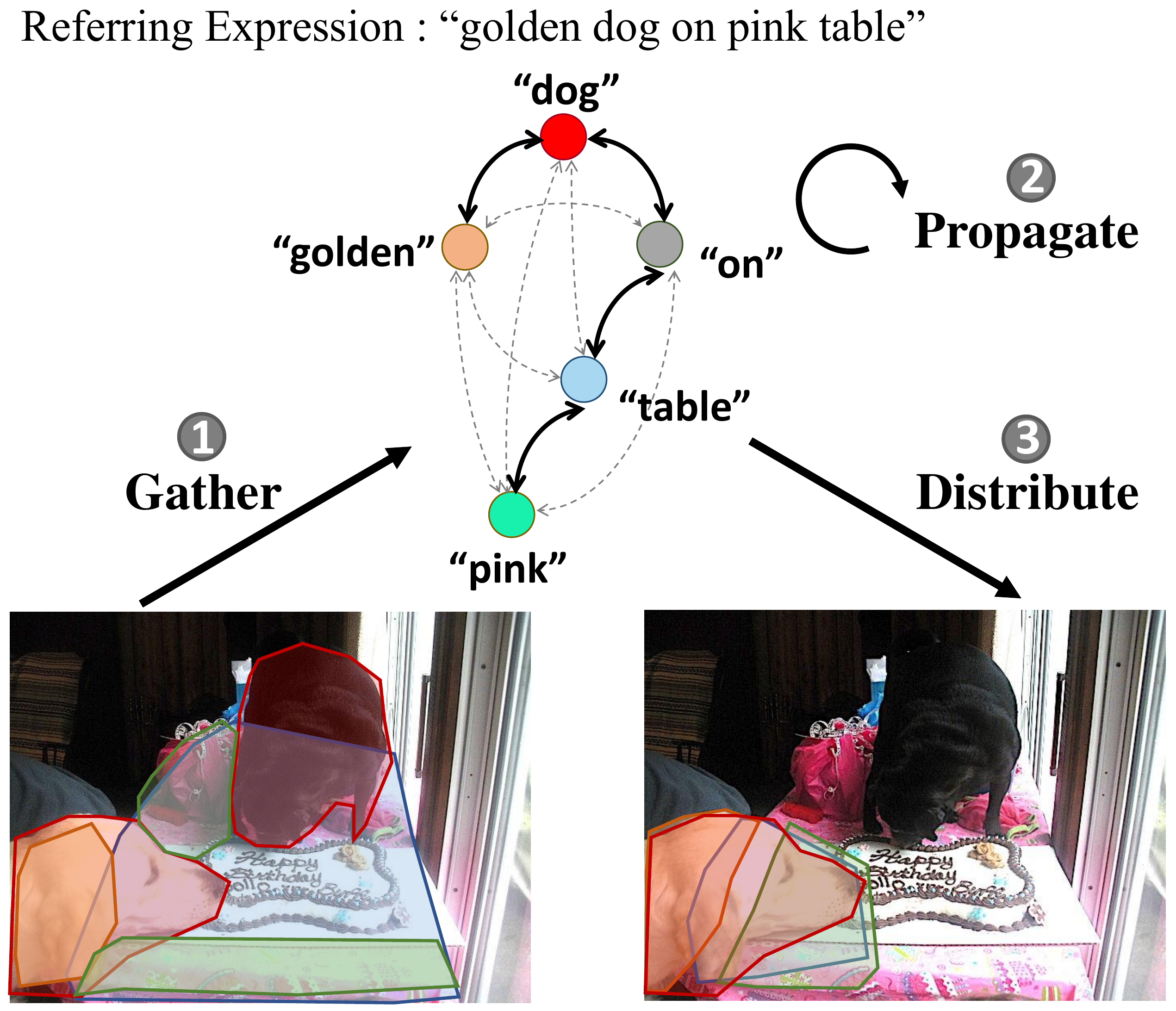}
    \caption{Illustration of our proposed LSCM module. 
    We construct a Dependency Parsing Tree suppressed Word Graph (DPT-WG) to model multimodal context in three steps.
    1) \textbf{Gather.} Multimodal context relevant to each word are gathered as feature of each word node. Therefore, each word corresponds to some visually relevant segments in the image. For example, word ``dog'' corresponds to two red segments in the left image. 
    2) \textbf{Propagate.} DPT is exploited to further guide each word node to include valid multimodal context from others and exclude disturbing ones through suppressed graph propagation routes. Gray dotted and black solid lines denote suppressed and unsuppressed edges in DPT-WG respectively. 
    3) \textbf{Distribute.} Features of all word nodes are distributed back to the image. Segments corresponding to the input words are all clustered around the ground-truth segmentation region, i.e., the golden dog on pink table in the right image. (Best viewed in color).}
	\label{fig:motivation}
\end{figure}

To obtain valid multimodal context, a feasible solution is to exploit linguistic structure as guidance to selectively model valid multimodal context which is relevant to the sentence. 
As illustrated in Fig.~\ref{fig:motivation}, each word can gather multimodal context related to itself by cross-modal attention. 
For example, the word ``dog'' corresponds to the red masks of two dogs in the image. 
Multimodal context of each word is a partial and isolated comprehension result of the whole sentence. 
Therefore, constrained communication among words is required to include valid multimodal context and exclude disturbing ones. 
Afterwards, communicated multimodal context of each word contains appropriate information relevant to the whole sentence and can be aggregated to form valid multimodal context for highlighting features of the referent. 

To realize the above solution, we propose a Linguistic Structure guided multimodal Context Modeling (LSCM) module in this paper. 
Concretely, features of the input sentence and image are first fused to form the multimodal features. 
Then, as illustrated in Fig.~\ref{fig:motivation}, in order to fully exploit the linguistic structure of the input sentence, we construct a Dependency Parsing Tree suppressed Word Graph (DPT-WG) where each node corresponds to a word. Based on the DPT-WG, three steps are conducted to model valid multimodal context of the sentence. 
(1) \textbf{Gather} relevant multimodal features (i.e., context) corresponding to a specific word through cross-modal attention as the node feature. At this step, each word node contains only multimodal context related to itself. Take Fig.~\ref{fig:motivation} as an example, the segments corresponding to ``dog'' and ``table'' are denoted by red and blue masks respectively. The multimodal features inside each mask are attentively gathered to form the node feature of the graph. 
(2) \textbf{Propagate} information among word nodes so that each word node can obtain multimodal context of the whole sentence. 
Initially, nodes in the word graph are fully-connected without any constraint on the edge weights.
However, two words in the sentence may not be closely relevant to each other and unconstrained communication between them may introduce disturbing multimodal context. 
For example, the words ``golden'' and ``pink'' in Fig.~\ref{fig:motivation} modify different 
entities respectively (``dog'' and ``table'') and have relatively weak relevance between each other. 
Unconstrained (i.e., extensive) information propagation between ``golden'' and ``pink'' is unnecessary and may introduce disturbing multimodal context. 
Therefore, we utilize Dependency Parsing Tree (DPT)~\cite{chen2014fast} to describe syntactic structures among words to selectively suppress certain weights of edges in our word graph. The DPT-WG can guide each word node to include valid contexts from others and exclude disturbing ones. 
After propagation, updated node features acquire information of the whole sentence. 
As shown in Fig.~\ref{fig:motivation}, the five words communicate and update their features under the structural guidance of our DPT-WG. 
(3) \textbf{Distribute} the updated node features back to every spatial location on the multimodal feature map. As shown in Fig.~\ref{fig:motivation}, the segments corresponding to the input words are all clustered around the ground-truth referring segmentation. It shows the updated multimodal features contain more valid multimodal context. 
In addition, we also propose a Dual-Path Multi-Level Fusion module which integrates spatial details of low-level features and semantic information of high-level features using bottom-up and top-down paths to refine segmentation results. 

The main contributions of our paper are summarized as follows: 
\begin{itemize}
  \item We introduce a ``gather-propagate-distribute'' scheme to model compact multimodal context by interaction between visual and linguistic modalities.
  \item We implement the above scheme by proposing a Linguistic Structure guided Context Modeling (LSCM) module which can aggregate valid multimodal context and exclude disturbing ones under the guidance of Dependency Parsing Tree suppressed Word Graph (DPT-WG). Thus, more discriminative multimodal features of the referent are obtained.
  \item Extensive experiments on four benchmarks demonstrate that our method outperforms all the previous state-of-the-arts, i.e., UNC ($+1.58\%$), UNC+ ($+3.09\%$), G-Ref ($+1.65\%$) and ReferIt ($+2.44\%$).
\end{itemize}

\section{Related Work}
\subsection{Semantic Segmentation}
In recent years, semantic segmentation has made great progress with Fully Convolutional Network~\cite{long2015fully} based methods.
DeepLab~\cite{chen2014semantic} replaces standard convolution with atrous convolution to enlarge the receptive field of filters, 
leading to larger feature maps with richer semantic information than original FCN. DeepLab v2~\cite{chen2017deeplab} and v3~\cite{chen2017rethinking} 
employ parallel atrous convolutions with different atrous rates called ASPP to aggregate multi-scale context. 
PSPNet~\cite{zhao2017pyramid} adopts a pyramid pooling module to capture multi-scale information. 
EncNet~\cite{zhang2018context} encodes semantic category prior information of the scenes to provide global context. 
Many works~\cite{badrinarayanan2017segnet}\cite{lin2017refinenet} exploit 
low level features  containing  detailed information to refine local parts of segmentation results.

\subsection{Referring Image Localization \& Segmentation}
Referring image localization aims to localize the object referred by a natural language expression with a bounding box. 
Some works~\cite{hu2017modeling}\cite{yang2019cross}\cite{liao2020real} model the relationships 
between multimodal features to match the objects with the expression. 
MAttNet~\cite{yu2018mattnet} decomposes the referring expression into subject, location and relationship to compute modular scores for localizing the referent. 
Comparing with referring image localization, referring image segmentation aims to obtain a more accurate result of the referred object, i.e., a semantic mask instead of a bounding box. 
Methods in the referring segmentation field can be divided into two types, i.e., bottom-up and top-down. 
\textbf{Bottom-up} methods mainly focus on multimodal feature fusion to directly predict the mask of the referent. Hu \textit{et al}~\cite{hu2016segmentation} proposes a straightforward concatenation of visual and linguistic features from CNN and LSTM~\cite{hochreiter1997long}. 
Multi-level feature fusion are exploited in~\cite{li2018referring}.
Word attention~\cite{shi2018key}\cite{chen2019see}, multimodal LSTM~\cite{liu2017recurrent}\cite{margffoy2018dynamic} and adversarial learning~\cite{qiu2019referring} are further incorporated to refine multimodal features. 
Cross-modal self-attention is exploited in~\cite{ye2019cross} to capture the 
long-range dependencies between image regions and words, introducing much redundant context due to the dense computation of self-attention. 
\textbf{Top-down} methods mainly rely on pretrained pixel-level detectors, i.e., Mask R-CNN~\cite{He2017MaskR} to generate RoI proposals and predict the mask within the selected proposal. 
MAttNet~\cite{yu2018mattnet} incorporates modular scores into Mask R-CNN framework to conduct referring segmentation task. Recent CAC~\cite{chen2019referring} introduces cycle-consistency between referring expression and its reconstructed caption into Mask R-CNN to boost the segmentation performance. 
In this paper, we propose a bottom-up method which exploits linguistic structure as guidance to include valid multimodal context and exclude disturbing ones for accurate referring segmentation.

\subsection{Structural Context Modeling}
Modeling context information is vital to vision and language problems. 
Typical methods like self-attention~\cite{vaswani2017attention}\cite{wang2018non} has shown great power for capturing the long range dependencies within the linguistic or visual modality.
In addition, more complicated data structures are also explored to model context information. 
Chen \textit{et al}~\cite{chen2019graph} proposes a latent graph with a small number of nodes to capture context from visual features for recognition and segmentation. In referring expression task, graphs~\cite{hu2019language}\cite{yang2019cross}\cite{yang2019dynamic}\cite{yang2020graph} using region proposals as nodes and neural module tree traversal~\cite{liu2019learning} are also explored to model multimodal contexts to some extent. Different from them, we propose to build a more compact graph using referring words as nodes and exploit dependency parsing tree~\cite{chen2014fast} to selectively model valid multimodal context.

\section{Method}
\begin{figure}[t]
  \centering
  \includegraphics[width=1.0\linewidth]{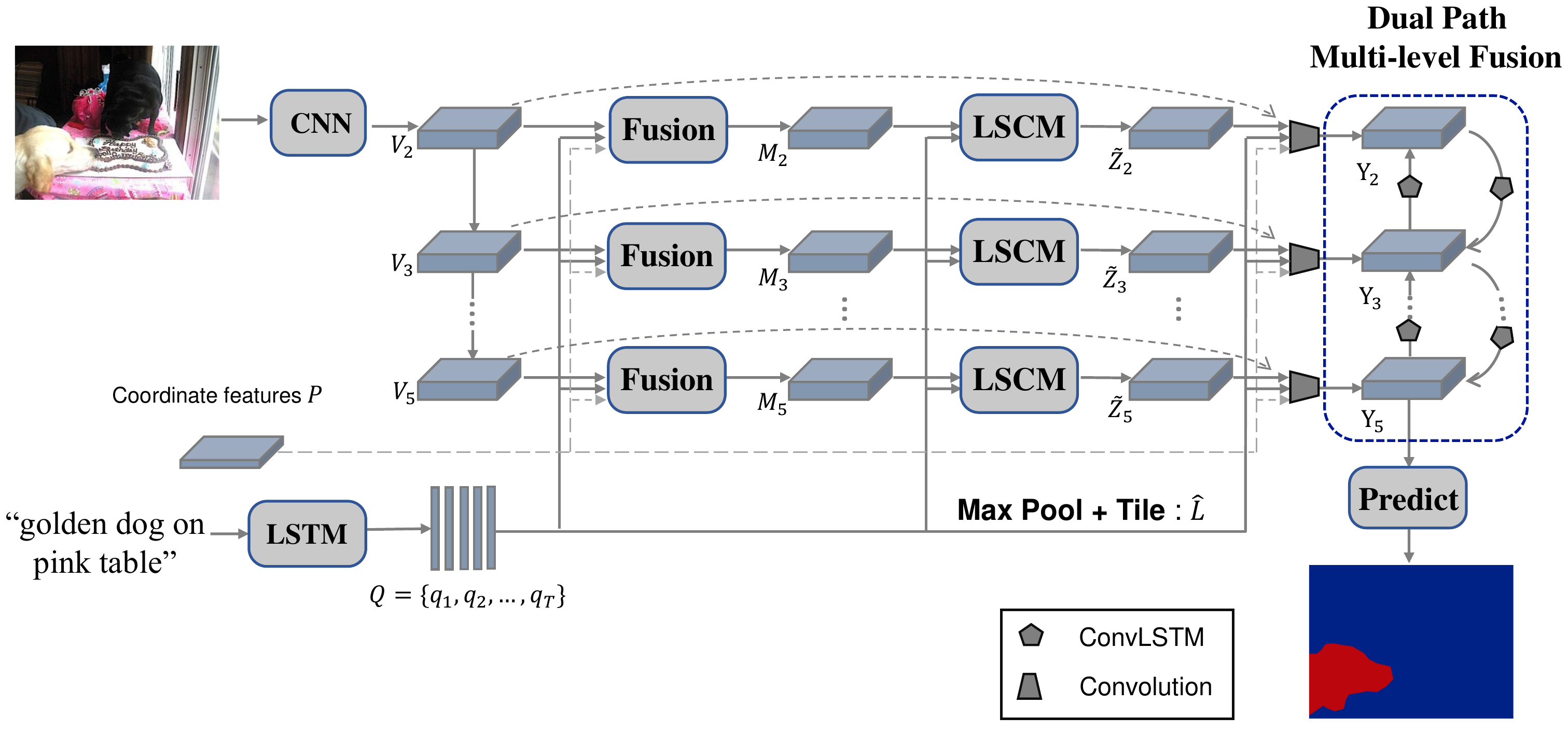}
  \caption{Overall architecture of our model. Multi-level visual features $V_i, i \in [2, 5]$, word features $Q$ and coordinate feature $P$ are first fused to get multimodal features $M_i$. Then $M_i$ are fed into our proposed LSCM to model valid multimodal context guided by linguistic structures. The output features $\tilde{Z}_i$ are combined with previous features and further fused through our Dual-Path Multi-Level Fusion module for mask prediction.}
  \label{fig:pipeline}
\end{figure}

The overall architecture of our model is illustrated in Fig.~\ref{fig:pipeline}. 
We first extract visual and linguistic features with a CNN and an LSTM respectively and then fuse them to obtain the multimodal feature. 
Afterwards, the multimodal feature is fed into our proposed Linguistic Structure guided Context Modeling (LSCM) module to highlight multimodal features of the referred entity. 
Our LSCM module conducts context modeling over the multimodal features under the structural guidance of DPT-WG.
Finally, multi-level features are fused by our proposed Dual-Path Fusion module for mask prediction.

\subsection{Multimodal Feature Extraction}
Our model takes an image and a referring sentence with $T$ words as input. 
As shown in Fig.~\ref{fig:pipeline}, we use a CNN backbone to extract multi-level visual features and then transform them to the same size. 
Multi-level visual features $\{V_2, V_3, V_4, V_5\}$ correspond to 
$\{Res2, Res3, Res4, Res5\}$ features of ResNet~\cite{he2016deep}, where $V_i \in \mathbb{R}^{H \times W \times C_v}, i\in \{2, 3, 4, 5\}$, with $H$, $W$ and $C_v$ being the height, width and channel number of visual features respectively. 
Since we conduct the same operations on each level of the visual features, we use $V$ to denote a single level of them for ease of presentation. 
For the input sentence of $T$ words, we generate features of all the words $Q \in \mathbb{R}^{T \times C_l}$ with an LSTM~\cite{hochreiter1997long}. 
To incorporate more spatial information, we also use an $8$D spatial coordinate feature~\cite{liu2017recurrent} denoted as $P \in \mathbb{R}^{H \times W \times 8}$. 
Afterwards, we fuse the features $\{V, Q, P\}$ to form the multimodal feature $M \in \mathbb{R}^{H \times W\times C_h}$, for which a simplified Mutan fusion~\cite{ben2017mutan} is adopted in this paper: $M = Mutan(V, Q, P)$. 
Details of Mutan fusion are included in the supplementary materials. 
Note that our method is not restricted to Mutan fusion, any other multimodal fusion approach can be used here.

\subsection{Linguistic Structure Guided Context Modeling}
\begin{figure}[t]
    \centering
    \includegraphics[width=1.0\linewidth]{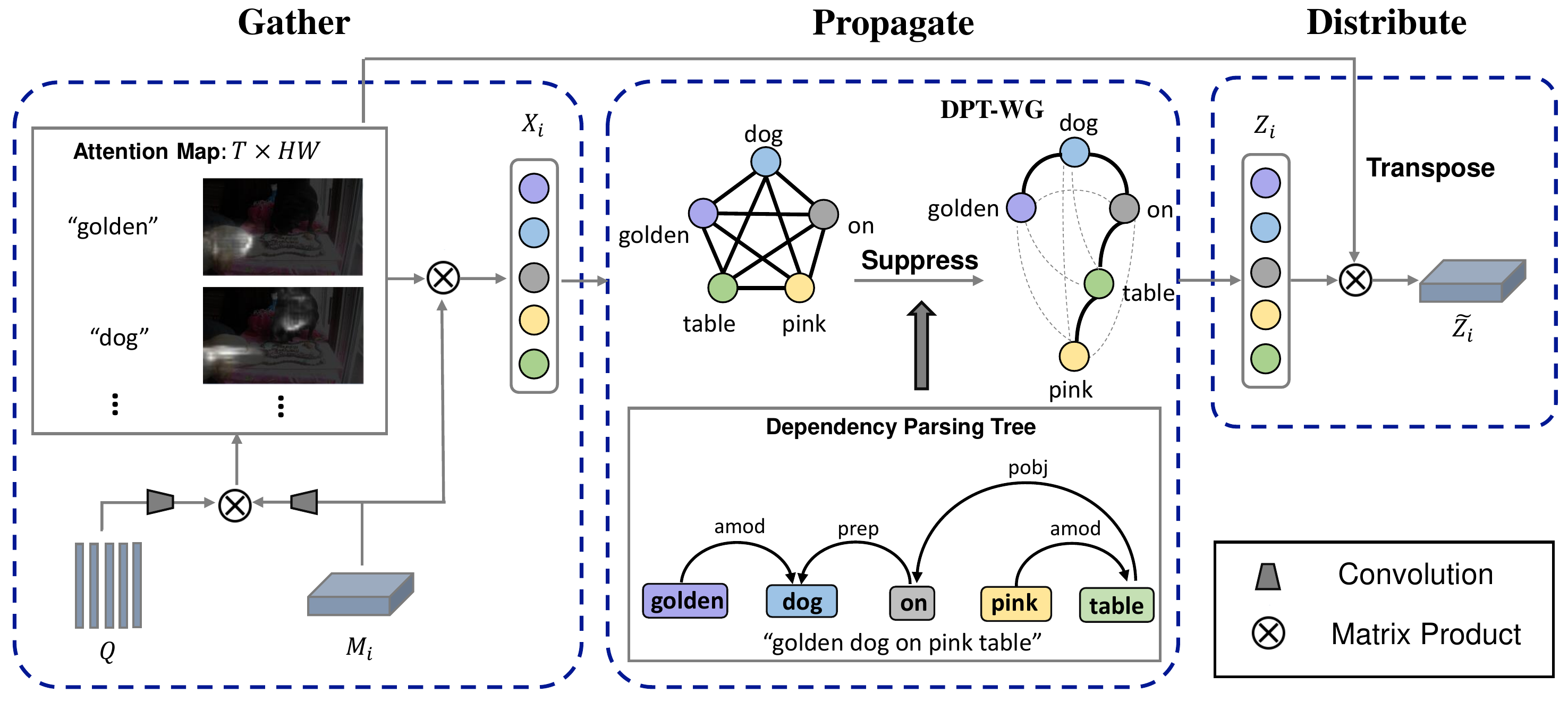}
    \caption{Illustration of our LSCM module. We use cross-modal attention between words features $Q$ and multimodal feature $M_i$ to gather feature for each word node. Then we exploit DPT to softly suppress disturbing edges in the initial fully-connected WG and conduct information propagation. Finally, the updated features of word nodes are distributed back as $\tilde{Z_i}$ to incorporate valid multimodal context into original features.}
    \label{fig:lscm}
\end{figure}

In this module, we build a Dependency Parsing Tree suppressed Word Graph (DPT-WG) to model valid multimodal context. 
As illustrated in Fig~\ref{fig:lscm}, we first gather feature vectors of all the spatial locations on multimodal feature $M$ into $T$ word nodes of WG. 
Then we exploit DPT~\cite{chen2014fast} to softly suppress the disturbing edges in WG for selectively propagating information among word nodes, which includes valid multimodal contexts while excluding disturbing ones. 
Finally, we distribute features of word nodes back to each spatial location.

\textbf{Gather}: 
We get a cross-modal attention map $B \in \mathbb{R}^{T \times HW}$ with necessary reshape and transpose operations as follows:
\begin{equation}
    \label{eq:slcr_fk_wq}
    B^{\prime} = (QW_{q2})(MW_m)^T,
\end{equation}
\begin{equation}
    \label{eq:slcr_norm_wg}
    B = Softmax(\frac{B^{\prime}}{\sqrt{C_h}}),
\end{equation}
where $W_{q2} \in \mathbb{R}^{C_l \times C_h}$ and $W_m \in \mathbb{R}^{C_h \times C_h}$ are learned parameters. 
Then we apply the normalized attention map $B$ to $M$ to gather the features into $T$ word nodes:
\begin{equation}
    \label{eq:slcr_proj}
    X = BM,
\end{equation}
where $X=[x_1; x_2; ...; x_T] \in \mathbb{R}^{T\times C_h}$ denotes the features of word nodes. 
Each $x_t, t = 1, 2, ..., T$ encodes the multimodal context related with the $t$-th word.

\textbf{Propagate}: 
The word graph used for context modeling is fully-connected. Thus, the adjacency matrix $A \in \mathbb{R}^{T\times T}$ is computed as follows:
\begin{equation}
    \label{eq:slcr_ag}
    A^{\prime} = (XW_{x1})(XW_{x2})^T,
\end{equation}
\begin{equation}
    \label{eq:slcr_norm_mat}
    A = Softmax(\frac{A^{\prime}}{\sqrt{C_h}}),
\end{equation}
where $W_{x1} \in \mathbb{R}^{C_h \times C_h}$, $W_{x2} \in \mathbb{R}^{C_h \times C_h}$ are parameters for linear transformation layers. 
At present, the edge weights among word nodes are represented by multimodal feature similarities which are unconstrained. 
However, two words may not be closely related in the sentence and unconstrained information propagation between them may introduce plenty of noises, yielding disturbing multimodal context. 
To alleviate this issue, we exploit DPT to selectively suppress disturbing edges which do not belong to the DPT structure. 
Concretely, we compute a tree mask $S \in \mathbb{R}^{T \times T}$ to restrict the adjacency matrix $A$ as follows: 
\begin{equation}
    \label{eq:tree_mask}
    S_{ij} = 
        \begin{cases}
            1, \; i \in \mathcal{C}(j) \; or \; j \in \mathcal{C}(i) \\
            \\
            \alpha, \; otherwise,
        \end{cases}
\end{equation}
where $i, j \in [1, T]$ are nodes in the parsing tree, $\mathcal{C}(j)$ denotes the children nodes set of node $j$, and $\alpha$ is a hyperparameter which is set as $0.1$ in our paper.
Then we multiply the adjacency matrix $A$ with the tree mask $S$ elementwisely to obtain a soft tree propagation route $A_t$ to diffuse information on the graph by:
\begin{equation}
    \label{eq:tree_a}
    A_t = A \odot S,
\end{equation}
where $\odot$ is elementwise multiplication. 
We then adopt one graph convolution layer~\cite{kipf2016semi} to propagate and update node features as follows:
\begin{equation}
    \label{eq:slcr_gc}
    Z  = (A_t + I)XW_z,
\end{equation}
where $I$ is an identity matrix serving as shortcut connection to ease optimization, $W_z \in \mathbb{R}^{C_h \times C_h}$ is 
the parameter for updating node features, and $Z \in \mathbb{R}^{T \times C_h}$ is the output of the graph convolution. 
After propagation, each word node can include valid multimodal context and exclude disturbing ones through the proper edges in parsing tree, forming robust features aligned with the whole sentence.

\textbf{Distribute}: 
Finally, we distribute the updated features of word graph nodes $Z$ back to all the spatial locations using the transpose of $B$ by:
\begin{equation}
    \label{eq:slcr_proj_back}
    \tilde{Z}  = B^TZ.
\end{equation}
We further conduct max pooling over word features $Q \in \mathbb{R}^{T \times C_l}$ to obtain sentence feature $L \in \mathbb{R}^{C_l}$, and then tile $L$ for $H \times W$ times to form grid-like sentence feature $\hat{L} \in \mathbb{R}^{H \times W \times C_l}$. 
As shown in Fig.~\ref{fig:pipeline}, the distributed feature $\tilde{Z} \in \mathbb{R}^{H \times W \times C_h}$ is concatenated with $V$, $\hat{L}$ and $P$ and then fed into a $1 \times 1$ convolution to get the output feature $Y \in \mathbb{R}^{H \times W \times C_o}$.

\subsection{Dual-Path Multi-Level Feature Fusion}
It has been shown that the integration of features at different levels can lead to significant performance improvement of referring image segmentation~\cite{li2018referring}\cite{ye2019cross}\cite{chen2019see}. 
We therefore also extract $4$ levels of visual features $\{V_2, V_3, V_4, V_5\}$ as the input of our LSCM module. 
Then we utilize convolutional LSTM~\cite{xingjian2015convolutional} to fuse the output features of the LSCM module $\{Y_2, Y_3, Y_4, Y_5\}$.
The fusion process is illustrated in Fig.~\ref{fig:pipeline}. We propose a Dual-Path Multi-Level Fusion module which sequentially
fuses the features from $4$ levels through the bottom-up and top-down paths. 
The input sequence of ConvLSTM is $[Y_5, Y_4, Y_3, Y_2, Y_3, Y_4, Y_5]$. 
The first bottom-up path sequentially integrates low-level features, which is able to complement high-level features with spatial details to refine the local parts of the mask. 
However, high-level features, which are critical for the model to recognize and localize the overall contour of the referred 
entities, are gradually diluted when integrating more and more low-level features. 
Thus, the top-down fusion path which reuses $Y_3$, $Y_4$ and $Y_5$ after bottom-up path is adopted to supplement more semantic multimodal information. 
Our Dual-Path Multi-Level Fusion module serves as a role to enhance features with both high-level semantics and low-level details for better segmentation performance.

\section{Experiments}

\subsection{Experimental Setting}

\textbf{Datasets}: 
We conduct extensive experiments on four benchmarks including UNC~\cite{yu2016modeling}, UNC+~\cite{yu2016modeling}, G-Ref~\cite{mao2016generation} 
and ReferIt~\cite{kazemzadeh2014referitgame}.
UNC and UNC+~\cite{yu2016modeling} are both collected from MS COCO dataset~\cite{lin2014microsoft}. 
The UNC dataset contains $19,994$ images with $142,209$ referring expressions for $50,000$ objects while the UNC+ dataset contains $19,992$ images with 
$141,564$ expressions for $49,856$ objects. UNC+ has no location words hence it is more challenging than UNC. 
G-Ref\cite{mao2016generation} is also built upon the MS COCO dataset~\cite{lin2014microsoft}. 
It consists of $26,711$ images with $104,560$ referring expressions for $54,822$ objects. The expressions are of average length of $8.4$ 
words which is much longer than that of the other three datasets (with average length less than $4$). 
ReferIt\cite{kazemzadeh2014referitgame} is composed of $19,894$ images with $130,525$ referring expressions for $96,654$ objects. It also contains stuff categories.

\textbf{Implementation Details}: 
Following previous works~\cite{li2018referring}\cite{ye2019cross}, we choose DeepLab-ResNet101~\cite{chen2017deeplab} pre-trained on Pascal VOC dataset~\cite{everingham2010pascal} as our backbone CNN. 
\textit{Res}$2$, \textit{Res}$3$, \textit{Res}$4$ and \textit{Res}$5$ are adopted for multi-level feature fusion. 
Input image is resized to $320 \times 320$. 
The maximum length of each referring expression is set to $20$. 
For feature dimensions, we set $C_v = C_l = C_h = 1000, C_o = 500$. 
$\alpha = 0.1$ in our final model. 
The network is trained using Adam optimizer~\cite{kingma2014adam} with an initial learning rate of $2.5e^{-4}$ and a weight decay of $5e^{-4}$. 
We apply a polynomial decay with power of $0.9$ to the learning rate. 
CNN is fixed during training. 
We use batch size $1$ and stop training after $700K$ iterations. 
GloVe word embeddings~\cite{pennington2014glove} pretrained on Common Crawl with $840B$ tokens are used to replace randomly initialized ones. 
For fair comparison with prior works, all the final segmentation results are refined by DenseCRF~\cite{krahenbuhl2011efficient}.

\textbf{Evaluation Metrics}: 
Following the setup of prior works~\cite{hu2016segmentation}\cite{li2018referring}\cite{ye2019cross}\cite{chen2019see}, we adopt overall intersection-over-union (\textit{Overall IoU}) and precision with different thresholds (\textit{Pr@X}) as the evaluation metrics for our model. 
The \textit{Overall IoU} is calculated by dividing the total intersection area with the total union area, where both intersection area and union area are accumulated over all test samples. 
The \textit{Pr@X} measures the percentage of prediction masks whose \textit{IoU} is higher than the threshold \textit{X}, where $\textit{X} \in \left\{ 0.5, 0.6, 0.7, 0.8, 0.9 \right\}$. 

\begin{table}[!htbp]
\centering
\caption{Comparison with state-of-the-art methods on four benchmarks using \textit{Overall IoU} as metric. ``n/a'' denotes methods does not use the same split as others. ``BU'' and ``TD'' denote ``Bottom-Up'' and ``Top-Down'' respectively.}
\small
\begin{tabular}{|c|c|c|c|ccc|ccc|c|}
\hline
\multirow{2}*{Type} & \multirow{2}*{Method} & & ReferIt & & UNC & & & UNC+ & & G-Ref \\
& & & test & val & testA & testB & val & testA & testB & val \\
\hline
\hline
\multirow{2}*{TD} & MAttNet~\cite{yu2018mattnet} & & - & 56.51 & 62.37 & 51.70 & 46.67 & 52.39 & 40.08 & n/a \\
& CAC~\cite{chen2019referring} & & - & 58.90 & 61.77 & 53.81 & - & - & - & 44.32 \\
& NMTree~\cite{liu2019learning} & & - & 56.59 & 63.02 & 52.06 & 47.40 & 53.01 & 41.56 & n/a \\
\hline
\hline
\multirow{9}*{BU} & LSTM-CNN~\cite{hu2016segmentation} & & 48.03 & - & - & - & - & - & - & 28.14 \\
& DMN~\cite{margffoy2018dynamic} & & 52.81 & 49.78 & 54.83 & 45.13 & 38.88 & 44.22 & 32.29 & 36.76 \\
& RMI~\cite{liu2017recurrent} & & 58.73 & 45.18 & 45.69 & 45.57 & 29.86 & 30.48 & 29.50 & 34.52 \\
& KWA~\cite{shi2018key} & & 59.09 & - & - & - & - & - & - & 36.92 \\
& CMSA(vgg16)~\cite{ye2019cross} & & 59.91 & 52.38 & 54.68 & 49.59 & 34.41 & 36.53 & 30.10 & 32.35 \\
& ASGN~\cite{qiu2019referring} & & 60.31 & 50.46 & 51.20 & 49.27 & 38.41 & 39.79 & 35.97 & 41.36 \\
& RRN~\cite{li2018referring} & & 63.63 & 55.33 & 57.26 & 53.95 & 39.75 & 42.15 & 36.11 & 36.45 \\
& Ours(vgg16) & & 63.82 & 55.41 & 57.92 & 52.54 & 41.18 & 44.32 & 35.78 & 39.78 \\
& CMSA~\cite{ye2019cross} & & 63.80 & 58.32 & 60.61 & 55.09 & 43.76 & 47.60 & 37.89 & 39.98 \\
& STEP~\cite{chen2019see} & & 64.13 & 60.04 & 63.46 & 57.97 & 48.19 & 52.33 & 40.41 & 46.40 \\
& Ours & & \textbf{66.57} & \textbf{61.47} & \textbf{64.99} & \textbf{59.55} & \textbf{49.34} & \textbf{53.12} & \textbf{43.50} & \textbf{48.05} \\
\hline
\end{tabular}
\label{tab:sota}
\end{table}

\subsection{Comparison with State-of-the-arts}
Table~\ref{tab:sota} summarizes the comparison results in \textit{Overall IoU} between our method and previous state-of-the-art methods. 
As illustrated in Table~\ref{tab:sota}, our method consistently outperforms both bottom-up and top-down state-of-the-art methods on four benchmark datasets. 

For bottom-up methods, STEP~\cite{chen2019see} densely fuses $5$ feature levels for $25$ times and achieves notable performance gains over CMSA~\cite{ye2019cross}. Our method outperforms STEP on all the splits using less times of multimodal feature fusion, which indicates that our LSCM can capture more valid mulitmodal context information to better align features between visual and linguistic modalities. Particularly, ReferIt is a challenging dataset on which pervious methods only achieve marginal improvements. CMSA and STEP outperform RRN~\cite{li2018referring} by $0.17\%$ and $0.50\%$ IoU respectively, while our method significantly boost the performance gain to $2.94\%$, which well demonstrates the effectiveness of our method. Moreover, on UNC+ dataset which has no location words, our method also achieves $3.09\%$ over STEP on testB split, showing that our method can model richer multimodal context information with less input conditions. In addition, we reimplement CSMA using their released code and our method using VGG16 as backbone. Our VGG16-based method also yields better performance on all $4$ datasets with margins of $3.24\%$ on UNC, $7.79\%$ on UNC+, $7.43\%$ on G-Ref and $3.91\%$ on ReferIt dataset, showing that our method can well adapt to different visual features.

For top-down methods, MAttNet~\cite{yu2018mattnet} and CAC~\cite{chen2019referring} first generate a set of object proposals and then predict the foreground mask within the selected proposal. The decoupling of detection and segmentation relies on Mask-RCNN which is pretrained on much more COCO images ($110K$) than bottom-up methods using only PASCAL-VOC images ($10K$) for pretraining. Therefore, comparing their performances with bottom-up methods may not be completely fair. However, our method still outperforms MAttNet and CAC with large margins, indicating the superiority of our method. In addition, on ReferIt dataset which contains sentences about stuff, our method achieves state-of-the-art performance while top-down methods may not be able to well handle them. 

There are also many top-down works~\cite{hu2019language}\cite{yang2019cross}\cite{yang2019dynamic}\cite{yang2020graph} in referring localization field which adopt graphs to conduct grounding. Their graphs are composed of region proposals which rely on detectors pretrained on COCO and/or other large datasets. However, our DPT-WG consists of referring words and uses DPT to suppress disturbing edges in WG. Then, features of WG are distributed back to highlight grid format features of the referent for bottom-up mask prediction. Thus, our method is also different from NMTree~\cite{liu2019learning} in which neural modules are assembled to tree nodes to conduct progressive grounding (i.e., retrieval) based on region proposals.

\subsection{Ablation Studies}
We perform ablation studies on UNC val set to verify the effectiveness of our proposed LSCM module and the Dual-Path Fusion module for leveraging multi-level features. Experimental results are summarized in Table~\ref{tab:ablation}. 

\begin{table}[t]
    \begin{center}
    \caption{Ablation studies on UNC val set. All models use the same backbone (DeepLab-ResNet101) and DenseCRF for postprocessing. 
    $^\ast$The statistics of RRN-CNN~\cite{li2018referring} are higher than those reported in the original paper which do not use DenseCRF. 
    Row $8$ and row $11$ are the same models with different names.}
    \begin{tabular}{|c|c|c|c|c|c|c|c|}
    \hline
    & Method & \textit{Pr@0.5} & \textit{Pr@0.6} & \textit{Pr@0.7} & \textit{Pr@0.8} & \textit{Pr@0.9} & \textit{IoU}(\%) \\
    \hline
    1 & RRN-CNN~\cite{li2018referring}$^\ast$ & 46.99 & 37.96 & 27.86 & 16.25 & 3.75 & 47.26 \\
    2 & +LSCM & 61.26 & 52.93 & 43.39 & 27.38 & 6.70 & 54.87 \\
    3 & +LSCM, GloVe~\cite{pennington2014glove} & 63.13 & 54.20 & 43.38 & 27.54 & 6.78 & 55.93 \\
    4 & +LSCM, GloVe, Mutan~\cite{ben2017mutan} & \textbf{64.25} & \textbf{55.64} & \textbf{45.00} & \textbf{29.24} & \textbf{7.28} & \textbf{56.50} \\
    \hline
    5 & Multi-Level-RRN-CNN~\cite{li2018referring}$^\ast$ & 65.83 & 57.45 & 46.76 & 31.91 & 10.40 & 57.61 \\
    6 & +LSCM & 68.33 & 61.16 & 51.59 & 36.98 & 11.57 & 59.67 \\
    7 & +LSCM, GloVe~\cite{pennington2014glove} & 70.56 & 62.89 & 52.91 & 38.07 & 11.99 & 60.98 \\
    8 & +LSCM, GloVe, Mutan~\cite{ben2017mutan} & \textbf{70.84} & \textbf{63.82} & \textbf{53.67} & \textbf{38.69} & \textbf{12.06} & \textbf{61.54} \\
    \hline
    9 & +Concat Fusion & 68.49 & 60.78 & 50.92 & 34.87 & 9.94 & 60.10 \\
    10 & +Gated Fusion~\cite{ye2019cross} & 69.08 & 62.46 & 50.73 & 35.42 & 11.27 & 60.46 \\
    11 & +Dual-Path Fusion (Ours) & \textbf{70.84} & \textbf{63.82} & \textbf{53.67} & \textbf{38.69} & \textbf{12.06} & \textbf{61.54} \\
    \hline
    \end{tabular}
    \label{tab:ablation}
    \end{center}
\end{table}

\textbf{LSCM module}: 
We first explore the effectiveness of our proposed LSCM module based on single level feature. 
Following~\cite{ye2019cross}, we implement the RRN-CNN~\cite{li2018referring} model without the recurrent refinement module as our baseline. 
Our baseline uses an LSTM to encode the whole referring expression as a sentence feature vector, and then concatenates it with each spatial location of the \textit{Res$5$} feature from DeepLab-101. 
Fusion and prediction are conducted on the concatenated features for generating final mask results. 
As shown in rows $1$ to $4$ of Table~\ref{tab:ablation}, \textbf{+LSCM} indicates that introducing our LSCM module into the baseline model can bring a significant performance gain of $7.61\%$ IoU, demonstrating that our LSCM can well model valid multimodal context under the guidance of linguistic structure.
Row $3$ and Row $4$ show that incorporating GloVe~\cite{pennington2014glove} and Mutan~\cite{ben2017mutan} fusion can further boost the performance based on our LSCM module. 

We further conduct the same ablation studies based on multi-level features. All the models use our proposed Dual-Path Fusion module to fuse multi-level features. 
As shown in rows $5$ to $8$ of Table~\ref{tab:ablation}, our multi-level models achieve consistent performance improvements as the single-level models. 
These results well prove that our LSCM module can effectively capture multi-level context as well.
Moreover, we additionally adapt \textbf{GloRe}~\cite{chen2019graph} for the referring segmentation task over multi-level features and achieve $58.53\%$ IoU and $67.23\%$ \textit{Pr@$0.5$}. The adapted GloRe uses learned projection matrix to project multimodal features into fixed number of abstract graph nodes, then conducts graph convolutions and reprojection to refine multimodal features. Our \textbf{+LSCM} in row $6$ outperforms GloRe by $1.14\%$ IoU and $1.10\%$ \textit{Pr@$0.5$}, indicating that building word graph by cross-modal attention and incorporating DPT to suppress disturbing edges between word nodes can better model valid multimodal context than GloRe.

\textbf{Multi-level feature fusion}: 
We compare different methods including Concat Fusion, Gated Fusion~\cite{ye2019cross} and our Dual-Path Fusion for multi-level feature fusion. 
All the fusion methods take $4$ levels of multimodal features processed by our LSCM module as input. 
As shown in rows $9$ to $11$ of Table~\ref{tab:ablation}, our proposed Dual-Path Multi-Level Fusion module achieves the best result, showing the effectiveness of integrating both high-level semantics and low-level details. 
In addition, the gated fusion from~\cite{ye2019cross} conducts $9$ fusion operations while ours conducts $6$ fusion operations with better performance.

\begin{table}[!htbp]
    \centering
    \caption{Experiments of graph convolution in terms of \textit{Overall IoU}. 
    $n$ denotes number of layers of graph convolution in our LSCM module. $\alpha = 0.1$ here.}
    \begin{tabular}{|c|c|c|c|c|c|c|}
        \hline
          & \multicolumn{6}{|c|}{+LSCM, GloVe} \\
        \cline{2-7}
         & $n = 0$ & $n = 1$ & $n = 2$ & $n = 3$ & $n = 4$ & adaptive \\
        \hline
        UNC val  & 54.51 & \textbf{55.93} & 50.77 & 50.64 & 49.59 & 54.69 \\
        G-Ref val & 38.94 & \textbf{40.54} & 39.29 & 37.74 & 37.50 & 37.41 \\
        \hline
    \end{tabular}
    \label{tab:gcn}
\end{table}

\textbf{Layers of graph convolution}: In Table~\ref{tab:gcn}, we explore the effects of conducting different layers of graph convolution in our LSCM module on UNC val set and G-Ref val set. 
The results show that the naive increase of graph convolution layers in LSCM will deteriorate the segmentation performance, probably because multiple rounds of message propagation among all words muddle the multimodal context of each word instead of enhancing it. 
Besides, adaptive which means number of the graph convolution layers equal to the depth of DPT, yields lower performance than one layer of graph convolution. It indicates that propagating information among word nodes without further constrain will include more disturbing context. 
Conducting $1$ layer of graph convolution to communication between parents and children nodes is already sufficient without introducing too much noises, which also makes our method more efficient.
In addition, $n = 1$ also outperforms $n = 0$ which shows communication among words is necessary after gathering multimodal context for each word.

\begin{table}[!htbp]
    \centering
    \caption{\textit{Overall IoU} results of different edge weights $\alpha$ in tree mask $S$. 
    Experiments are conducted on UNC val set. All the models use $n = 1$ layer of graph convolution.}
    \begin{tabular}{|c|c|c|c|c|c|c|}
        \hline
        \multicolumn{7}{|c|}{+LSCM, GloVe} \\
        \hline
          $\alpha = 0$ & $\alpha = 0.1$ & $\alpha = 0.2$ & $\alpha = 0.3$ & $\alpha = 0.4$ & $\alpha = 0.5$ & $\alpha = 1$ \\
        \hline
        55.01 & \textbf{55.93} & 55.44 & 55.49 & 54.77 & 55.12 & 54.81 \\
        \hline
    \end{tabular}
    \label{tab:alpha}
\end{table}

\textbf{Edge weights in tree mask}: In Table~\ref{tab:alpha}, we explore how different values of $\alpha$ in the tree mask $S$ (Eq~\ref{eq:tree_mask}) influence the performance of our LSCM. We can observe that $\alpha = 0.1$ achieves the best performance and outperforms $\alpha = 1$ (WG w/o DPT) by $1.12\%$ IoU, which demonstrates that suppressing syntactic irrelevant edges in our word graph can reduce unnecessary information propagation and exclude disturbing multimodal context. In addition, $\alpha = 0$ yields inferior performance to $\alpha = 0.1$, indicating our DPT-WG (i.e., approximate spanning tree) can obtain more sufficient information than a strict DPT.

\begin{figure}[!htbp]
    \centering
    \includegraphics[width=0.9\linewidth]{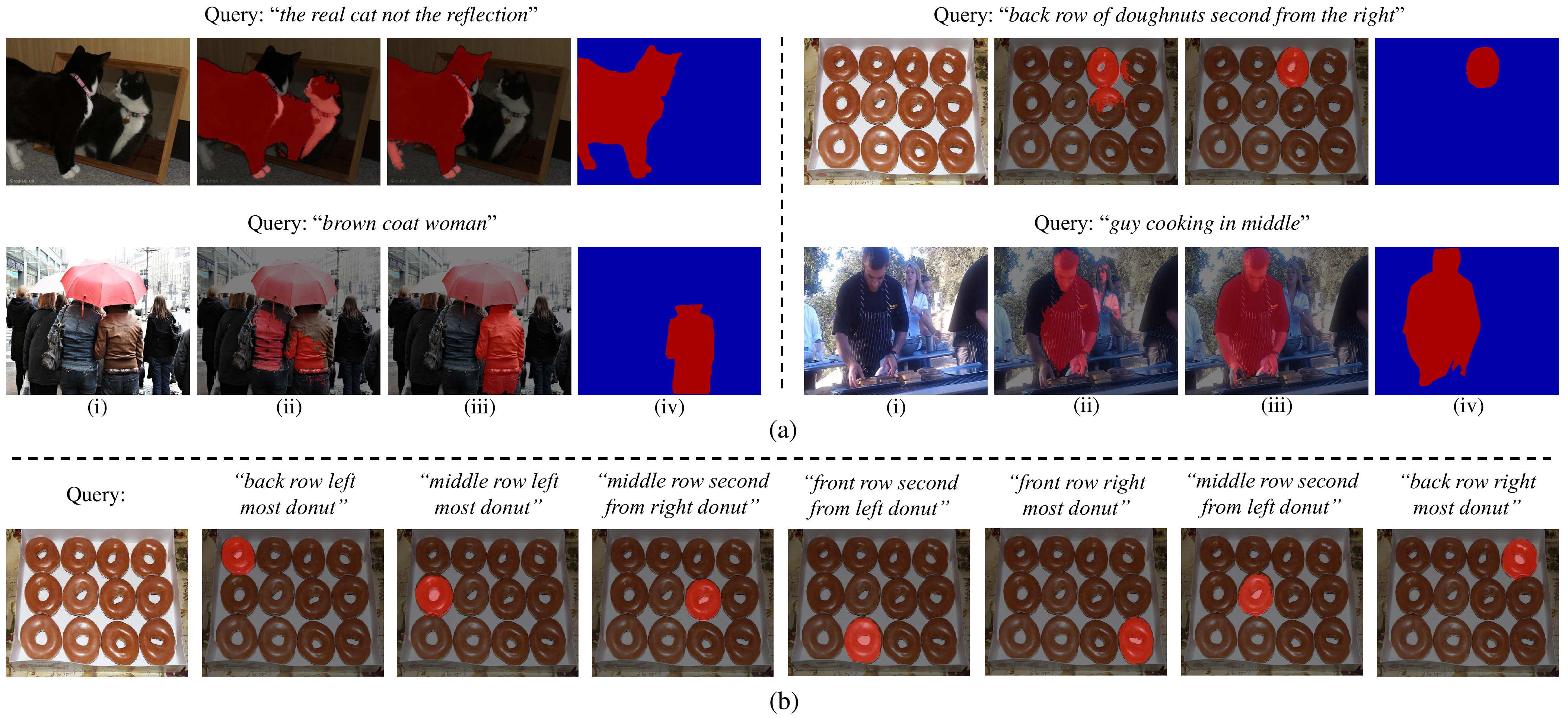}
    \caption{Qualitative Results of referring image segmentation. (a)(i) Original image. 
    (a)(ii) Results produced by the multi-level RRN-CNN baseline (row $5$ in Table~\ref{tab:ablation}). 
    (a)(iii) Results produced by our full model (row $8$ in Table~\ref{tab:ablation}). 
    (a)(iv) Ground-truth.
    (b) Results of customized expressions. Our model can adapt to new expressions flexibly.}
    \label{fig:qualitative}
\end{figure}

\textbf{Qualitative results}: 
Fig.~\ref{fig:qualitative}(a) presents the segmentation results predicted by our full model (row $8$ in Table~\ref{tab:ablation}) and the multi-level RRN-CNN baseline (row $5$ in Table~\ref{tab:ablation}). Comparing (b) and (c) in Fig.~\ref{fig:qualitative}, we can find that only multi-level feature refinement without valid multimodal context modeling is not sufficient for the model to understand the referring expression comprehensively, thus resulting in inaccurate predictions, such as segmenting ``coat'' but ignoring ``brown'' in the bottom-left of Fig.~\ref{fig:qualitative}. 
As shown in Fig.~\ref{fig:qualitative}(b), we also manually generate customized expressions to traverse many the donuts. 
It is interesting to find that our model can always understand different expressions adaptively and locate the right donuts, indicating that our model is flexible and controllable. 
More qualitative results on four datasets are presented in supplementary materials.

\begin{figure}[!htbp]
    \centering
    \includegraphics[width=0.83\linewidth]{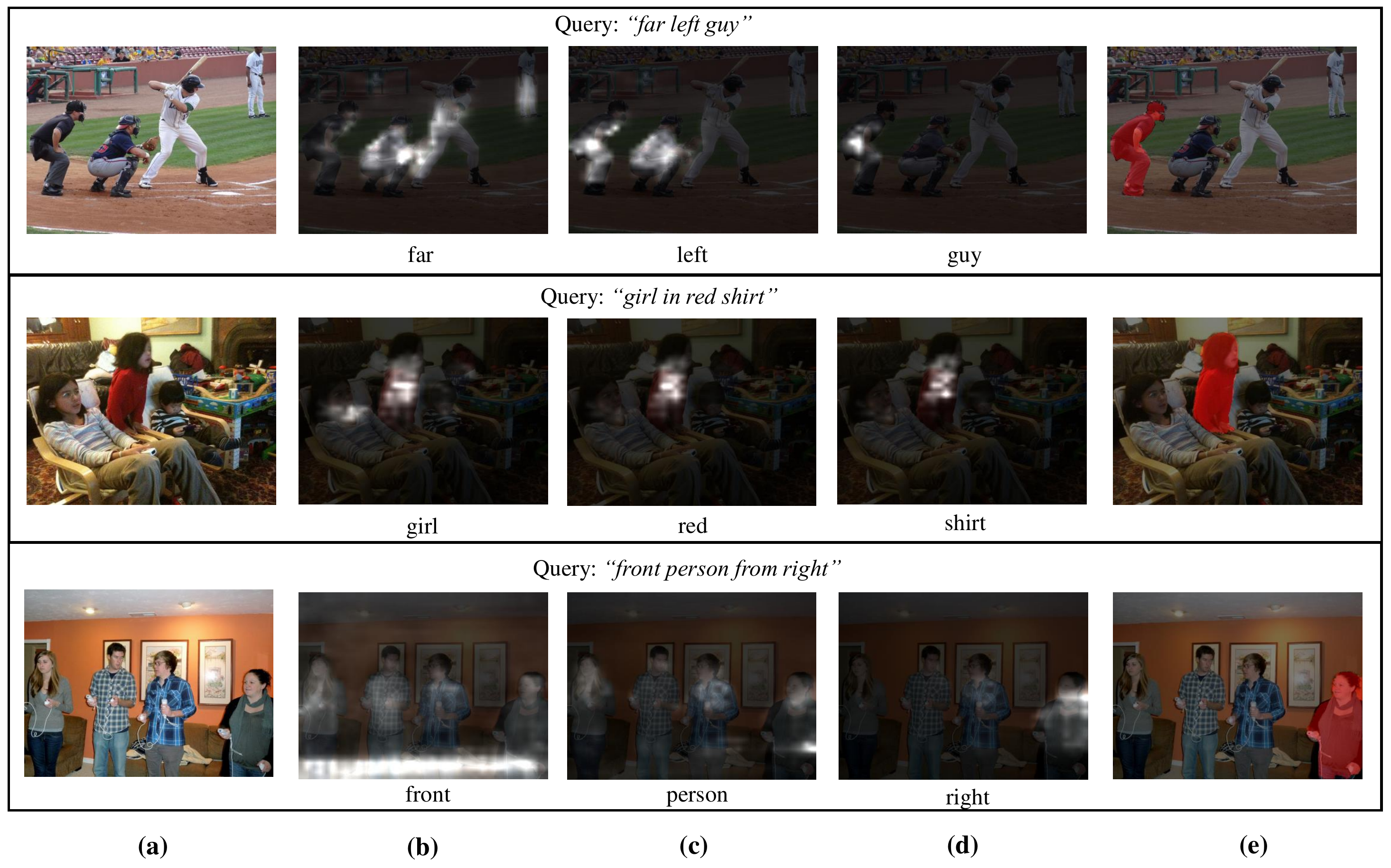}
    \caption{Visualization of attention maps on the given words. (a) the original image. 
    (b)(c)(d) refer to the attention maps of the specific words below. (e) predictions of our proposed method.}
    \label{fig:vis_attn}
\end{figure}

\textbf{Visualization of attention maps}: 
To give a straightforward explanation about how our LSCM works, we visualize the attention maps of each node (corresponding to the words of referring expression) to the spatial locations and the results are shown in Fig~\ref{fig:vis_attn}. 
The cross-modal attention maps correspond to $B$ obtained in the gather operation (Eq.~\ref{eq:slcr_fk_wq} and \ref{eq:slcr_norm_wg}), which has size of $T \times HW$. 
Each row of $B$ denotes the attention map of a certain word. 
The three words are organized in sequential order. 
From Fig.~\ref{fig:vis_attn} we find that a meaningful word usually attends to its corresponding area in the image. 
For example, in the third row of (b), the word ``front'' attends to the front area of the image, and in the second row of (c), word ``red'' attends to the area of red shirt. 
Our LSCM module is able to model valid multimodal context among these attended areas to obtain a precise segmentation of the referring expression.

\section{Conclusion and Future Work}
In this paper, we explore the referring image segmentation problem by introducing a ``gather-propagate-distribute'' scheme to model multimodal context. We implement this scheme as a Linguistic Structure guided Context Modeling (LSCM) module. 
Our LSCM builds a Dependency Parsing Tree suppressed Word Graph (DPT-WG) which guides all the words to include valid multimodal context of the sentence while excluding disturbing ones, which can effectively highlight multimodal features of the referent. 
Our proposed model achieves state-of-the-art performance on four benchmarks. 
In the future, we plan to adapt our LSCM module into other tasks (e.g., VQA, Captioning) to verify its effectiveness.

\subsubsection{Acknowledgement:} 
This work is supported by Guangdong Basic and Applied Basic Research Foundation (No. 2020B1515020048), National Natural Science Foundation of China (Grant 61876177, Grant 61976250), Beijing Natural Science Foundation (L182013, 4202034), Fundamental Research Funds for the Central Universities, Zhejiang Lab (No. 2019KD0AB04) and Tencent Open Fund.

\clearpage
%
%
\bibliographystyle{splncs04}
\bibliography{egbib}
\end{document}